%% file: SymFlux-arXiv.tex
\documentclass{gtpart}
\usepackage[utf8]{inputenc} 
\usepackage[T1]{fontenc}    
\usepackage{hyperref}       

\hypersetup{
  colorlinks   = true, 
  urlcolor     = teal, 
  linkcolor    = teal, 
  citecolor   = teal 
}

\usepackage{url}            
\usepackage{booktabs}       
\usepackage{amsfonts}       
\usepackage{nicefrac}       
\usepackage{microtype}      
\usepackage{cleveref}

\usepackage{amsmath,amsfonts}
\usepackage{algorithmic}
\usepackage{algorithm}
\usepackage{array}
\usepackage[caption=false,font=normalsize,labelfont=sf,textfont=sf]{subfig}
\usepackage{textcomp}
\usepackage{stfloats}
\usepackage{url}
\usepackage{verbatim}
\usepackage{graphicx}
\usepackage{cite}

\usepackage{amsthm}
\usepackage{multicol}
\usepackage{threeparttable}

\usepackage{booktabs}
\usepackage{hyperref}

\usepackage{color}
    \usepackage{xcolor, colortbl}

\usepackage[color=teal!70!white, bordercolor=teal, textsize=footnotesize,obeyFinal]{todonotes}
\setuptodonotes{inline}

\newtheorem{definition}{Definition}[section]
\newtheorem{Theorem}{Theorem}
\newtheorem{theorem}[Theorem]{Theorem}

\definecolor{gray}{rgb}{.85, .85, .85}

\begin{document}
\title{SymFlux: deep symbolic regression \\ of Hamiltonian vector fields}

\author{M. Evangelista-Alvarado}
\address{Universidad Nacional Rosario Castellanos (UNRC), Mexico City, Mexico.}
\email{miguel.eva.alv\,@rcastellanos.cdmx.gob.mx}

\author{P. Su\'arez-Serrato}
\address{Max-Planck Institute for Mathematics, Bonn, and Electrical and Computer Engineering, UC Santa Barbara, California, USA, and Instituto de Matem\'aticas, Universidad Nacional Aut\'onoma de M\'exico (UNAM), Mexico Tenochtitlan, Mexico.}

\email{pablo\,@im.unam.mx}

\input{sec/0_abstract}

\maketitle

\input{sec/1_better_intro}
\input{sec/2_examples}
\input{sec/3_database}
 \input{sec/4_models}
\input{sec/5_conclusions}

\appendix
\input{sec/A1_appendix}


\bibliographystyle{plain}
\bibliography{references}

\vfill

\end{document}

%% file: sec/0_abstract.tex
\begin{abstract}
 
 We present \textit{SymFlux}, a novel deep learning framework that performs symbolic regression to identify Hamiltonian functions from their corresponding vector fields on the standard symplectic plane. 
 \textit{SymFlux} models utilize hybrid CNN-LSTM architectures to learn and output the symbolic mathematical expression of the underlying Hamiltonian. 
 Training and validation are conducted on newly developed datasets of Hamiltonian vector fields, a key contribution of this work.
 Our results demonstrate the model's effectiveness in accurately recovering these symbolic expressions, advancing automated discovery in Hamiltonian mechanics.
%
\end{abstract}

%% file: sec/1_better_intro.tex
\section{Introduction}\label{sec:intro}

Hamiltonian dynamics, a cornerstone of mathematics and physics, offers a concise and powerful framework for describing complex systems where the total energy remains constant. 
This formalism is essential for understanding conservative dynamics in diverse scientific domains. 
At its heart, Hamiltonian dynamics characterizes a system's evolution through its energy function, the Hamiltonian $H$ \cite{marsden02}. 
This function encapsulates the system's energy and also its underlying symmetries and structural properties, proving instrumental in advancing our comprehension of phenomena ranging from celestial mechanics to subatomic particle interactions.
This framework's elegance stimulated the development of efficient numerical methods, enabling researchers to simulate and analyze otherwise intractable complex systems \cite{ArnoldEDO92}.

Despite the descriptive power of Hamiltonian mechanics, a significant challenge lies in identifying or discovering the specific Hamiltonian function that governs an observed dynamical system. 
This inverse problem, known as the Hamiltonization problem, is crucial for applying the full analytical and predictive capabilities of the Hamiltonian framework \cite{ArnoldMeth89}. 
Addressing this challenge motivates the exploration of advanced data-driven techniques.

Symbolic regression (SR) emerges as a compelling machine learning paradigm for such discovery tasks. 
Unlike traditional regression methods that fit data to predefined model structures, SR aims to uncover the underlying mathematical expressions that describe a system's behavior directly from data \cite{Schmidt09}. 
In the context of dynamical systems, SR can identify the governing algebraic or differential equations. More broadly, symbolic discovery seeks to uncover hidden patterns and symbolic representations within complex datasets \cite{Makke24}.
These approaches yield models that are not only predictive but also interpretable, generalizable, and reflective of the system's intrinsic mechanisms. 
This potential for uncovering fundamental laws makes symbolic regression a transformative tool for scientific inquiry across fields like physics, engineering, economics, and biology \cite{koza94,Cranmer20, Silviu20, Zhengdao20, Lopez23}.
Furthermore, by automating the discovery of mathematical relationships, SR can democratize scientific exploration, empowering researchers to decipher complex data patterns without necessarily requiring extensive prior mathematical intuition.

To formally define Hamiltonian systems, we first introduce some essential concepts from differential geometry.
Let $M$ be a smooth manifold. 
A differential form $\omega$ on $M$ is \textit{closed} if its exterior derivative vanishes, i.e., $d\omega=0$. A form is \textit{non-degenerate} if the induced map from the tangent space $T_{p}M$ to its dual $T_{p}^{*}M$ is an isomorphism for all $p\in M$.
A \textit{symplectic manifold} $(M, \omega)$ is a smooth manifold $M$ equipped with a symplectic form $\omega$, which is a closed and non-degenerate 2-form.
A \textit{symplectomorphism} is a diffeomorphism $\phi: (M, \omega) \to (\bar{M}, \bar{\omega})$ between symplectic manifolds such that $\phi^* \bar{\omega} = \omega$, where $\phi^*$ denotes the pullback by $\phi$ \cite{mcduff98}.

Given a symplectic manifold $(M, \omega)$ and a smooth function $H:M\to\mathbf{R}$ (the Hamiltonian), there exists a unique vector field $X_{H}$ on $M$, called the \textit{Hamiltonian vector field}, defined by the relation:
\begin{equation}\label{eq:Hmiltonian_VF}
  \iota_{X_{H}}\omega = dH.
\end{equation}
This equation implies Hamilton's equations of motion: $\frac{\partial H}{\partial q_{j}}=-{\dot {p_{j}}}$ and $\frac{\partial H}{\partial p_{j}}={\dot {q_{j}}}$, for $j=1,\cdots, n$. 
Here, $q_{j}$ represent the generalized position coordinates and $p_{j}$ represent the generalized momenta, which together define the coordinates in the phase space of the system. The triple $(M, \omega, H)$ constitutes a \textit{Hamiltonian system} \cite{marsden02}.

The \textit{Hamiltonization problem} for a dynamical system on a smooth manifold $M$ (potentially equipped with a symplectic form $\omega$) involves determining whether its flow $\varphi$ can be described by a Hamiltonian function $H$. 
That is, can we find an $H$ such that $\varphi$ is the flow generated by the Hamiltonian vector field $X_H$?
\cite{ArnoldMeth89}.
A specific class of such systems are \textit{Lie--Hamilton systems}, which are Hamiltonian systems defined on a symplectic manifold admitting a Lie group action that preserves the Hamiltonian function. This inherent symmetry simplifies the analysis of the system's dynamics.
Ballesteros et al. ~\cite{ballesteros15} have provided a classification of Lie--Hamilton systems in the standard symplectic plane ($\mathbf{R}^2$ with its canonical symplectic form).

\begin{figure}[!ht]
    \centering
    \includegraphics[width=0.98\textwidth]{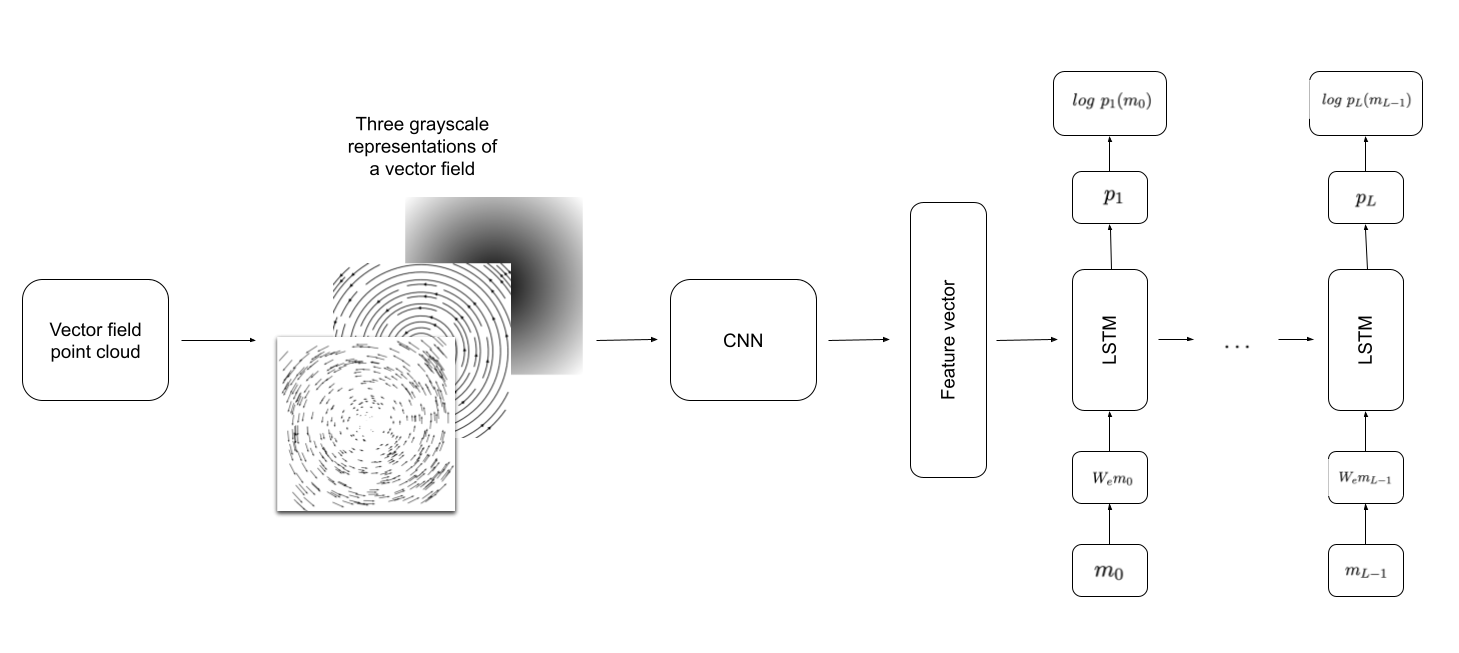} 
    \caption{A diagram of a general {\it SymFlux} system, employing CNN and LSTM submodules. This model processes an image representation of a vector field through a CNN to extract a feature vector. This vector is then input to an LSTM, which predicts the symbolic form of the Hamiltonian, $m_0 + \ldots + m_n$.}
    \label{fig:symflux}
\end{figure}

In this work, we address the Hamiltonization problem by employing symbolic regression powered by deep learning methodologies. 
Our central aim is to automatically discover the Hamiltonian function $H$ from observational data, specifically from visual representations of Hamiltonian vector fields. We introduce novel hybrid models, termed {\it SymFlux} systems (illustrated in \cref{fig:symflux}), which integrate Convolutional Neural Networks (CNNs)~\cite{Lecun98} for feature extraction from these visual inputs, and Long Short-Term Memory (LSTM) networks~\cite{Hochreiter97} for generating the symbolic representation of $H$.

Developing these models presented several challenges, primarily the scarcity of suitable datasets and the absence of directly comparable benchmark models for this specific task of symbolic Hamiltonization from visual data (see \cref{sec:related_work} for a discussion of related but distinct approaches). 
To overcome the data challenge, we undertook a systematic, multi-stage process:
\begin{enumerate}
    \item We first curated a diverse set of analytical functions to serve as ground-truth Hamiltonian energy functions with respect to the canonical symplectic structure in $\mathbf{R}^2$.
    \item Using these energy functions, we generated the corresponding symbolic and numerical databases of Hamiltonian vector fields.
    \item Finally, we created a database of visual representations (images) of these Hamiltonian vector fields, where each image is parameterized by its underlying energy function and a point-cloud sampling (detailed in \cref{sec:DB}).
\end{enumerate}
These three novel databases are crucial contributions that enable the training and evaluation of our deep learning architectures.

Our approach, inspired by image captioning techniques~\cite{vinyals15}, leverages the CNN component to process the visual representations of Hamiltonian vector fields and extract salient feature vectors. 
This CNN preprocessing also serves as an effective dimensionality reduction step, transforming the input from the four-dimensional tangent space where the vector fields reside to a more compact representation suitable for the subsequent symbolic regression task. 
The LSTM component then takes these feature vectors as input and, conditioned on our corpus of energy functions, infers the symbolic form of the Hamiltonian that generated the original vector field.

We developed and evaluated three distinct {\it SymFlux} models. 
Two of these use transfer learning, incorporating pre-trained Xception and ResNet architectures as the CNN submodule, while the third employs a CNN trained from scratch. 
In all such experimental configurations, the LSTM submodule was trained specifically for this task.

Our primary finding is that our {\it SymFlux} systems achieve a symbolic regression accuracy of 85\% to 88\% in identifying the correct Hamiltonian function from its visual representation (see \cref{sec:symflux} for detailed results). 
This demonstrates the viability of using deep learning for symbolic Hamiltonization, opening new avenues for data-driven symbolic discovery in dynamical systems.

\subsection{Related work}\label{sec:related_work}

The Hamiltonization problem has a rich theoretical background. 
The Lie-K\"onigs theorem~\cite{Whittaker88} provides a local solution for certain systems of ordinary differential equations, establishing conditions under which a local Lie transformation can convert such a system into an integrable Hamiltonian system. 
Hojman~\cite{Hojman96} addressed the Hamiltonization problem by adapting a Poisson structure to vector fields with an infinitesimal symmetry and a first integral.
Avenda\~no-Camacho et al.~\cite{Avendano22} extended Hojman's construction to vector fields with a transversally invariant metric, thereby solving the problem for low-dimensional torus Lie group actions.
Perlick~\cite{Perlick92} generalized the Lie--Königs theorem to even dimensions and allowed for time-dependent symplectic structures.

In recent years, machine learning techniques have been increasingly applied to problems in Hamiltonian dynamics. 
L\'opez--Pastor and Marquardt~\cite{Lopez23} developed a self-learning physics machine for inferences on time-reversible Hamiltonian systems, though unavailability of their source code prevents direct comparison. 
Cranmer et al.~\cite{Cranmer20} employed Graph Neural Networks (GNNs) to learn symbolic models from interacting particle systems. 
Greydanus et al.~\cite{Greydanus19} introduced a model that learns conservation laws in an unsupervised manner. 
Zhengdao et al.~\cite{Zhengdao20} used Recurrent Neural Networks (RNNs) to learn the dynamics of physical systems from observed trajectories. 
Transformer architectures have also been applied to symbolic inference tasks by Vastl et al.~\cite{Vastl22}. 
Other contributions to symbolic regression include the \textit{eureqa} software by Schmidt and Lipson~\cite{Schmidt09}, which discovers equations from data, and the \textit{AI Feynman} software by Udrescu and Tegmark~\cite{Silviu20} for symbolic regression of certain dynamical systems. 
Zhou et al.~\cite{Zhou19} developed an algorithm for numerically computing and visualizing Morse vector fields.

Previous research has significantly advanced the application of machine learning to dynamical systems and symbolic regression. 
However, no prior work has considered the symbolic regression of Hamiltonian vector fields, especially from visual data, as we do here. 
Adapting existing solutions for our specific task would be challenging, making direct performance comparisons difficult. 
Our work, therefore, pioneers a new direction in this interdisciplinary area.

\subsubsection{Contributions}

Our main contributions are:
\begin{enumerate}
    \item We introduce {\it SymFlux}, a novel deep learning framework for symbolic regression of Hamiltonian functions from visual representations of their corresponding vector fields on $\mathbf{R}^{2}$ with the standard symplectic structure.
    \item We present the creation of three new (synthetic) databases: a corpus of Hamiltonian functions, a database of their corresponding symbolic and numerical Hamiltonian vector fields, and a database of visual representations of these vector fields. These resources are specifically designed for training and evaluating machine learning models on the symbolic Hamiltonization task.
    \item Our databases systematically include all planar Lie--Hamiltonian vector fields that are not expressed as integrals, based on the classification by Ballesteros et al.~\cite{ballesteros15}.
    \item We demonstrate the efficacy of our approach, with {\it SymFlux} models achieving high accuracy in recovering the symbolic form of Hamiltonian functions, as summarized in \cref{tab:cnn_test_results}.
\end{enumerate}

\subsubsection{Organization}
The remainder of this article is organized as follows. 
In \cref{sec:Ham}, we present examples of Hamiltonian systems and detail our workflow for generating visual representations. \Cref{sec:DB}  describes the construction of our novel databases of Hamiltonian vector fields and their visual counterparts. 
\Cref{sec:symflux}  explains the architecture, training methodology, and experimental results of our {\it SymFlux} models. 
Finally, \cref{sec:conclusions} offers concluding remarks and discusses potential avenues for future research. \Cref{app:A} includes a datasheet for our datasets.

\subsection*{Acknowledgements}
The authors thank DGTIC-UNAM for access to UNAM's Miztli supercomputer HPC resources (grant LANCAD-UNAM-DGTIC-430), which were instrumental for training and experimenting with the deep learning models in this work. 
PSS thanks the Geometric Intelligence Laboratory at UC Santa Barbara for their hospitality and stimulating environment during Fall 2024, and the organizers of the AI in Mathematics and Theoretical Computer Science meeting at the Simons Institute for the Theory of Computing at UC Berkeley (Spring 2025) for fostering valuable interactions.
MEA acknowledges CONACyT for doctoral fellowship support during this research.

%% file: sec/2_examples.tex
\section{Examples of Hamiltonian systems}\label{sec:Ham}
The Darboux Theorem is crucial in symplectic geometry because it shows that all symplectic manifolds are locally indistinguishable (for a proof see \cite{mcduff98}). 

\begin{theorem}[Darboux Theorem]\label{teo:darboux}
In a symplectic manifold $(M, \omega)$, any point has a neighborhood that is symplectomorphic to $\mathbf{R}^{2n}$ with the standard symplectic form $$ \omega_0 = dx_1 \wedge dy_1 + \cdots + dx_n \wedge dy_n.$$ 
\end{theorem}

\subsection{Examples}\label{sec:HamExamples}

In this section, we will review some classical examples of Hamiltonian systems. 
The computations shown here, illustrated in figures \ref{fig:HarmonicExa} and \ref{fig:Math-Pendulumn}, were performed using our \texttt{poisson geometry} \cite{EPS1, EPS3} and \texttt{numerical poisson geometry} \cite{EPS2, EPS3} Python modules.

\subsubsection{$1$-dimensional harmonic oscillator}\label{sec:HarmonicExa}
The behavior of a mass on a spring, and molecule vibrations are phenomena modeled by a 1-dimensional harmonic oscillator, which describes a fundamental yet straightforward physical system \cite{ArnoldEDO92}.
The following second-order differential equation models the harmonic oscillator:
\begin{equation}\label{eqn:Harmonic-Oscillatorsecond}
    \ddot{x} + \alpha^{2} x = 0, \,\, x\in{\bf R} 
\end{equation}
Here $\alpha$ is a positive constant. Let $y=\dot{x}$, then \cref{eqn:Harmonic-Oscillatorsecond} transforms into the following system of first-order, linear, differential equations,
\begin{equation}\label{eqn:Harmonic-Oscillator}
    \dot{x}=y, \quad \quad \dot{y}=-\alpha^{2}x.
\end{equation} 
Observe that the system defined by  \cref{eqn:Harmonic-Oscillator} is Hamiltonian. The harmonic oscillator Hamiltonian is
\begin{equation}\label{eq:HarminicHamil}
H(x,y) = \frac{1}{2}\left( y^2 + \alpha^{2}x^{2} \right) .
\end{equation}
As we have an explicit formula for the Hamiltonian function, it is possible to obtain a symbolic expression for the vector field $X_{H}$ of the Hamiltonian function $H$:  
\begin{equation}\label{eq:HarmonicHVF}
    X_{H} = -y \frac{\partial}{\partial x} + x \frac{\partial}{\partial y}    
\end{equation}
Because the Hamiltonian function in \cref{eq:HarminicHamil} is a first integral of the one-dimensional harmonic oscillator, the trajectories of the system are contained within the level sets of $H$. 
Moreover, for every positive real number $k>0$, the level set $C_{k}=\left\{ (x,y)\, |\, y^{2}+\alpha^{2}x^{2}=2k \right\} $ of \cref{eq:HarminicHamil} is a regular closed curve. Consequently, the trajectories of the one-dimensional harmonic oscillator are precisely the ellipses describing the regular level sets of the function $H$. 
\begin{figure}[!ht]
    \includegraphics[width=0.98\textwidth]{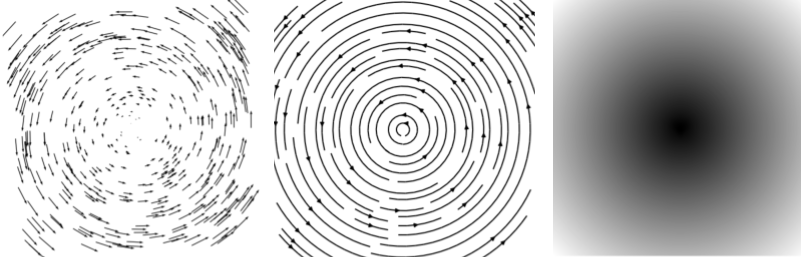}
    \caption{
    Three distinct images representing the harmonic oscillator Hamiltonian vector field in \cref{eq:HarmonicHVF}. For the case $\alpha=1$, the trajectories are circles describing the level set of the Hamiltonian function $H$ in \cref{eq:HarminicHamil}. We calculate the images (left: quiver method, center: streamline method, right: heatmap method) using our numerical Poisson geometry module \cite{EPS2} and  Python visualization tools.
    }
    \label{fig:HarmonicExa}
    \centering
\end{figure}

In \cref{sec:symflux} we show the results of the symbolic regression models developed in here, including for the $1$-dimensional harmonic oscillator, using the visual representations in \cref{fig:HarmonicExa} as input.

\subsubsection{Mathematical pendulum}\label{sec:PendulumExa}
A mathematical pendulum (or simple pendulum), is used to model the motion of a mass suspended from a fixed point by a string or rod.  
This example is frequently used to introduce concepts related to oscillations, conservation of energy, and harmonic motion \cite{ArnoldEDO92}.
The following system of differential equations describes the mathematical pendulum:
\begin{equation}\label{eqn:Math-Pendulumn}
\dot{x}= -\sin (y), \,\, \dot{y} = x
\end{equation}
Notice that this system of equations has one degree of freedom, and it is Hamiltonian with respect to the following function:
\begin{equation}\label{eqn:Math-PendulumnHamiltonian}
H(x,y) = \frac{1}{2}x^2 + \cos (y)
\end{equation}
Having the Hamiltonian function $H$ allows us to derive a precise, closed-form, description of the vector field  $X_{H}$:
\begin{equation}\label{eq:PendulumH_x}
    X_{H} = \sin (y) \frac{\partial}{\partial x} + x \frac{\partial}{\partial y}   
\end{equation}
Observe that the Hamiltonian function in \cref{eqn:Math-PendulumnHamiltonian} is a first integral of the mathematical pendulum, because the trajectories of this system correspond exactly to the curves representing the level sets of \cref{eqn:Math-PendulumnHamiltonian},  
$$C_{k}=\left\{ (x,y): \frac{1}{2}x^2 + \cos(y)= k \right\}.$$ 

\begin{figure}[!ht]
    \includegraphics[width=0.98\textwidth]{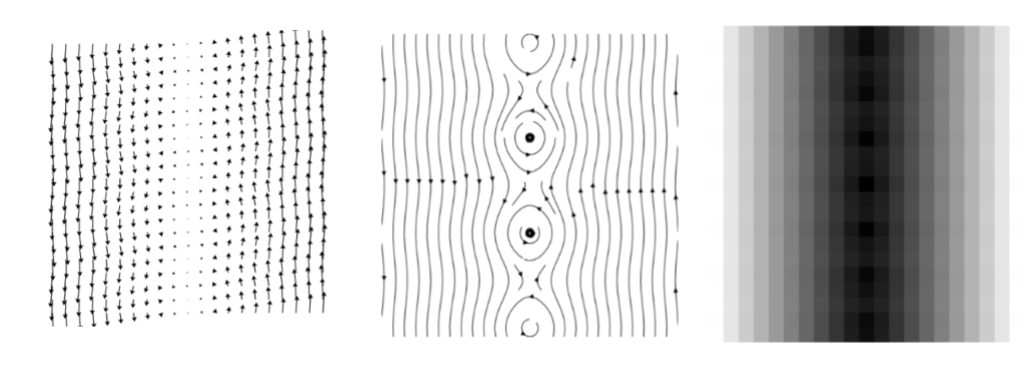}
    \caption{Three images associated with the mathematical pendulum's Hamiltonian vector field $X_{H}$ from \cref{eq:PendulumH_x}. We compute these images (left: quiver method, center: streamline method, right: heatmap method) using our Poisson geometry numerical module \cite{EPS2} and Python visualization modules.
    }
    \label{fig:Math-Pendulumn}
    \centering
\end{figure}

The vector field $X_{H}$ in \cref{eq:PendulumH_x} is illustrated in \cref{fig:Math-Pendulumn}.

In \cref{sec:symflux} we present the results of our symbolic regression models evaluated on the mathematical pendulum, based on the visual representations in \cref{fig:Math-Pendulumn}.

\subsubsection{Lotka--Volterra predator-prey model}\label{sec:Lotka-Volterra-ex}

\begin{figure}[!ht]
    \includegraphics[width=0.98\textwidth]{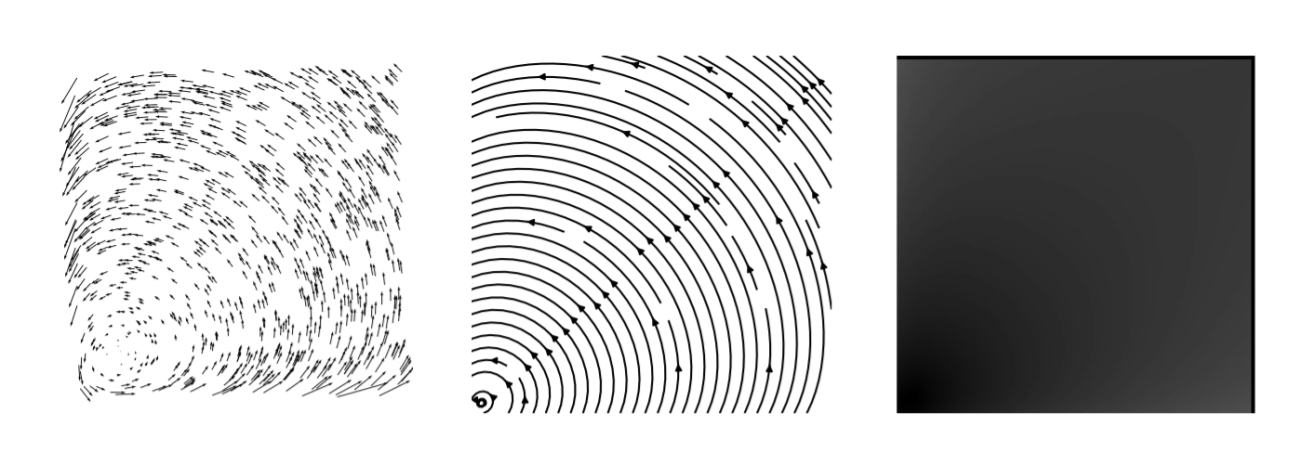}
    \caption{Three images associated with the Lotka--Volterra predator-prey model's Hamiltonian vector field $X_{H}$ from \cref{eqn:HamilVectFiledPredator-prey}. We compute these images (left: quiver method, center: streamline method, right: heatmap method) using our Poisson geometry numerical module \cite{EPS2} and Python visualization modules.
    }
    \label{fig:Predator-prey}
    \centering
\end{figure}

The Lotka--Volterra system, a model used to describe predator-prey interactions, can be expressed in Hamiltonian form using the following Hamiltonian function:
\begin{equation}\label{eqn:HamilPredator-prey}
H(x, y) = x \ln (x) + y \ln (y) - ax - by - cxy
\end{equation}
Here, $x$ represents the population of the prey species, and $y$ the population of the predator species.
The positive constants $a, b$, and $c$ represent the growth rate of the prey, the death rate of the predator, and the interaction rate between the two species, respectively \cite{lotka1925}.
The corresponding Hamiltonian equations of motion are:
\begin{equation}\label{eqn:Predator-prey}
\dot{x} = y - cxy, \quad\quad \dot{y} = -x + cx
\end{equation}
These equations are equivalent to the original Lotka--Volterra equations, so the system can be formulated as a Hamiltonian system.
Consider, as a specific example, the case where the constants are $a = b = 1.1$ and $c = 0.1$.
The Hamiltonian vector field associated with $H$ is then:
\begin{equation}\label{eqn:HamilVectFiledPredator-prey}
X_{H}= ( 0.1x - \log(y) + 0.1 ) \frac{\partial}{\partial x} + (\log(x) -0.1y - 0.1)\frac{\partial}{\partial y}
\end{equation}
\Cref{fig:Predator-prey} presents three visual representations of the vector field \ref{eqn:HamilVectFiledPredator-prey}.

In \cref{sec:symflux} we present the results of our symbolic regression models evaluated on the Lotka--Volterra system. 

\subsubsection{Epidemiological Hamiltonian Susceptible-Infectious-Susceptible Model}\label{sec:SISHamiltonian}
The spread of an infectious disease and its transmission dynamics may be modeled with the Susceptible-Infectious-Susceptible (SIS) model, which describes a fundamental system in epidemiology \cite{esen22}. 
In this model, two population groups are considered: the susceptible $S$ and the infected $I$. 
The dynamics of the SIS model are represented by a system of first-order differential equations. 
\begin{equation}\label{eqn:sis_system}
    \dot{x} = x\rho_{0} - x^{2} - \frac{1}{y^{2}}, \quad\quad \dot{y} = -y\rho_{0} + 2xy
\end{equation}
The coefficient $\rho_0$ in \cref{eqn:sis_system} is related to the epidemiological reproduction number, which estimates the number of new infections.
The system defined by \cref{eqn:sis_system} is Hamiltonian, its Hamiltonian function $H$ and its Hamiltonian vector field $X_{H}$ are:
\begin{equation}\label{eq:SISHamil}
H(x,y) = xy(\rho_{0}-x) + \frac{1}{y}
\end{equation}
\begin{equation}\label{eq:SISHVF}
    X_{H} = \left( -\rho_{0}x + x^{2} + y^{-2} \right) \frac{\partial}{\partial x} + y(\rho_{0} - 2x) \frac{\partial}{\partial y}    
\end{equation}

In \cref{sec:symflux} we present the results of our symbolic regression models evaluated on the Hamiltonian SIS system.

%% file: sec/3_database.tex
\section{Constructing databases of Hamiltonian Vector fields}\label{sec:DB}

This section describes the methods for constructing new Hamiltonian vector field datasets in two dimensions. 
Recall that by Darboux's \cref{teo:darboux} states that all symplectic manifolds of the same dimension are locally equivalent; that is, in a sufficiently small neighborhood, any symplectic 2-form can be transformed into the standard symplectic form.
This justifies our choice to investigate Hamiltonian vector fields on the standard symplectic plane, as all other symplectic structures are locally equivalent to it.

Additionally, by the Weierstrass approximation theorem \cite{Schenkman72}, any continuous function defined on a closed real interval can be uniformly approximated (arbitrarily well) by some polynomial function.
Therefore, by equation \cref{eq:Hmiltonian_VF}, we may approximate a Hamiltonian vector field through a parameterization based on polynomial energy functions.
This gives us a schematic way to generate an expressive family of Hamiltonian vector fields.

\subsection{Defining a basis of Hamiltonian functions}\label{subsec:functions}
During the construction of our databases, we must restrict ourselves to a finite number of energy functions.
We limit the complexity of the energy functions by selecting a finite set of representative polynomial functions of bounded degree. 
This method ensures that the database remains computationally manageable while maintaining a diverse range of Hamiltonian vector fields.
This includes the Lie--Hamiltonian vector fields (with respect to $\omega_0$), having the same degree bound and that is not expressed as an integral \cite{ballesteros15}.

\subsubsection{Polynomial energy functions}\label{subsub:function_polynomial}

Let $x,y\in \mathbf{R}$, $i$ be an integer greater than or equal to one, and $\Delta$ be a finite set of values.
Define the set $B_{i}(\Delta)$ as:
$$B_{i}(\Delta) := \left\{ b_{j}x^{h}y^{k} \,\, | \,\, 1 \leq h + k \leq i \text{ , } b_{j} \in \Delta, \text{ and } j, h, \text{and }  k \in \mathbf{N}\cup\{0\}  \right\}$$
Here, $\Delta$ represents all the coefficients that a polynomial can take. 
Observe that we can not take $\Delta$ to be the set of real numbers, since that would imply that each element $b_{j}\in\Delta$ would have an uncountable number of options, which is computationally impossible.
In other words, every $b_{j}x^{h}y^{k}\in B_{i}$ with $b_{j} \in \mathbf{R}$ would require infinite computational resources.

\subsubsection{Trigonometric energy functions}\label{subsub:function_trigo}

In order to introduce greater diversity and precisely recover the known symbolic modelds for a wide array of physical phenomena, we add trigonometric functions to $B_i(\Delta)$. 
Many physical systems, particularly those involving periodic behavior, are naturally described by trigonometric terms. 
For example, the mathematical pendulum in \cref{sec:PendulumExa}.
Let $T$ be a set defined as
$$T:=\{b_{j}t_{l} \text{ }|\text{ } t_{l}\in\{\cos(x), \cos(y), \sin(x), \sin(y)\} \text{ and }  b_{j} \in\Delta\}.$$ 
Define $B_{i}^{*}(\Delta):=B_{i}(\Delta) \cup T$; this combination allows us to model both polynomial and periodic behaviors within the same framework.

\subsubsection{Cardinality of energy function sets}\label{subsubsec:hamiltonian_set}

The cardinality of the set $B_i(\Delta)$ is determined by the possible values that the coefficients $b_j$ can take for each element $x^{h}y^{k}$.
That is, it is parameterized by the choice of the set  $\Delta$ and by $i$. The same applies to the set $B_{i}^{*}(\Delta)$.
Define the linear span of $B_{i}(\Delta)$ as 
$${\rm Span}(B_{i}(\Delta)) := \{\Sigma b_{j}x^{h}y^{k} \,\, | \,\, b_{j}x^{h}y^{k} \in B_{i}(\Delta)
\}.$$
Analogously, we can define the span of $B_{i}^{*}(\Delta)$ as
$${\rm Span}(B^{*}_{i}(\Delta)) := \{\Sigma\,b_{j} g_{l}\,\, | \,\, b_{j} g_{l}\in B^{*}_{i}(\Delta)
\}.$$
The sets ${\rm Span}(B_{i}(\Delta))$ and ${\rm Span}(B^{*}_{i}(\Delta))$ define two databases of energy functions by considering the linear combinations of the elements in $B_i$ and $B^{*}_{i}$, in the same order.
These spans provide a structured and comprehensive foundation for our databases.
Consider the following sets:
\begin{align*} 
\Delta_{3} &= \{-1, \textit{ } 0, \textit{ } 1\} \textit{,  
 } \quad\quad \Delta_{5} = \{-1, \textit{ } -0.5, \textit{ } 0, \textit{ } 0.5, \textit{ } 1\} \\ 
\Delta_{7} &= \{-1, \textit{ } -2/3 \textit{ } -1/3, \textit{ } 0, \textit{ } 1/3, \textit{ } 2/3, \textit{ } 1\} \textit{, and}\\
\Delta_{9} &= \{-1, \textit{ } -0.75, \textit{ } -0.5, \textit{ } -0.25, \textit{ } 0, \textit{ } 0.25, \textit{ } 0.5, \textit{ } 0.75, \textit{ } 1\}
\end{align*}
The coefficients $b_{j}$ are uniformly distributed on the interval $[-1,1]$, and the cardinality of $\Delta_{l}$ is equal to $l$. 
Define the set:
$${\rm Ham}(B_{i}, \Delta_{l}):=\{H \, | \, H\in {\rm Span}(B_{i}(\Delta_{l})), \text{ and } H\notin \mathbf{R}\}$$
Analogously, define ${\rm Ham}(B_{i}^{*}, \Delta_{l})$.

We need to remove the constant polynomials from the set ${\rm Ham}(B_{i}, \Delta_{l})$ because these generate trivial Hamiltonian vector fields. 
The associated mechanical system has no evolution in phase space and remains in a static state.

To clarify the descriptions above, we calculate the cardinality of the set ${\rm Ham}(B_{1}, \Delta_{3})$, which is \\
$\{-1x, 0x, 1x, -1y, 0y, 1y\}$. 
We obtain all the polynomials that can be constructed with such elements $\{-x-y,\,\, -x,\,\, -x + y,\,\, -y,\,\, 0,\,\, y,\,\, x-y,\,\, x,\,\, x + y\}$; that is, we calculate all the combinations of the elements containing the variable $x$ with the components of the variable $y$. 
The result is nine; we have three options for the variable $x$ and three for the variable $y$. 
Now, subtract the zero polynomial (generated by $0x + 0y$), the unique constant polynomial generated in this process. 
Finally, obtain the total cardinality of the set ${\rm Ham}(B_{1}, \Delta_{3})$, that is eight.
\Cref{tab:basisCard} shows the cardinality of all functions spanned by the sets ${\rm Ham}(B_{i}, \Delta_{l})$, with  $i \in \{1,2,3,4,5\}$ and $l \in \{3,5,7,9\}$.
\begin{table}[ht!]
    \centering
    \begin{tabular}{|c|c|c|c|c|c|}
        \hline
        step & $B_{1}$ & $B_{2}$ & $B_{3}$ & $B_{4}$ & $B_{5}$  \\
        \hline	
        $\Delta_{3}$ & $8$ & \cellcolor{gray}$242$ & \cellcolor{gray}$19,682$ & $4,782,968$ & $\pgfmathprintnumber{3486784400}$ \\
         
        $\Delta_{5}$ & $24$ &\cellcolor{gray} $3,213$ &  $1,953,124$ & $\pgfmathprintnumber{6103515624}$ & $\pgfmathprintnumber{95367431640624}$ \\

        $\Delta_{7}$ & $48$ & $16,806$ & $40,353,606$ &  $\pgfmathprintnumber{678223072848}$ & $\pgfmathprintnumber{79792266297612000}$ \\
        
        $\Delta_{9}$ & $80$ & $59,040$ & $\pgfmathprintnumber{387420488}$ & $\pgfmathprintnumber{22876792454960}$ & $\pgfmathprintnumber{12157665459056928800}$ \\
        
        \hline
        \end{tabular}
    \caption{Cardinalities of the sets ${\rm Ham}(B_{i}, \Delta_{l})$ for $i \in \{1,2,3,4,5\}$ and $l \in \{3,5,7,9\}$.}
    \label{tab:basisCard}
\end{table}

\Cref{tab:basisTrigoCard} shows the cardinalities of all functions spanned by the sets ${\rm Ham}(B_{i}^{*}, \Delta_{l})$, with $i \in \{1,2,3,4,5\}$ and $l \in \{3,5,7,9\}$.
\begin{table}[h!]
    \centering
    \begin{tabular}{|c|c|c|c|c|c|}
        \hline
        step & $B_{1}^{*}$ & $B_{2}^{*}$ & $B_{3}^{*}$ & $B_{4}^{*}$ & $B_{5}^{*}$  \\
        \hline				
        $\Delta_{3}$ & $729$ &  \cellcolor{gray}$19,683$ &  $1,594,323$ & $\pgfmathprintnumber{387420489}$ & $\pgfmathprintnumber{282429536481}$ \\
        
        $\Delta_{5}$ & $15,625$ & $1,953,125$ & $\pgfmathprintnumber{1220703125}$ & $\pgfmathprintnumber{3814697265625}$ & $\pgfmathprintnumber{59604644775390625}$ \\

        $\Delta_{7}$ & $117,649$ & $40,353,607$ & $\pgfmathprintnumber{96889010407}$ &  $\pgfmathprintnumber{1628413597910449}$ & $\pgfmathprintnumber{191581231380566414401}$ \\
        		
        $\Delta_{9}$ & $531,441$ & $\pgfmathprintnumber{387420489}$ & $\pgfmathprintnumber{2541865828329}$ & $\pgfmathprintnumber{150094635296999121}$ & $\pgfmathprintnumber{79766443076872509863361}$ \\
        		
        	
        \hline
        \end{tabular}
    \caption{Cardinalities of the sets ${\rm Ham}(B_{i}^{*}, \Delta_{l})$ for $l \in \{1,2,3,4,5\}$ and $j \in \{3,5,7,9\}$.}
    \label{tab:basisTrigoCard}
\end{table}

Therefore, to keep our computations tractable and within a reasonable computational size, here we will work with the function sets  ${\rm Ham}(B_2, \Delta_{3})$, ${\rm Ham}(B_2, \Delta_{5})$, ${\rm Ham}(B_3, \Delta_{3})$, and ${\rm Ham}(B_{2}^{*}, \Delta_{5})$. 
In \cref{tab:basisCard} and \cref{tab:basisTrigoCard} the cardinalities of these specific sets appear in gray.

\subsection{Symbolic database for Hamiltonian vector fiels}\label{subsec:db_symbolic}

We can now generate a broad variety of Hamiltonian systems parameterized by the energy functions ${\rm Ham}(B, \Delta)$ explained above. 
Let $n$ be the cardinality of ${\rm Ham}(B, \Delta)$. 
Define the symbolic dataset $\mathcal{X}_{s}(B, \Delta)$ as 
$\mathcal{X}_{s}(B, \Delta) := \{(H_i, X_{H_i}, \omega_{0}) \text{ }|\text{ } H_{i}\in{\rm Ham}(B, \Delta) \text{ and } \omega_{0} ={\rm d} x\wedge{\rm d}y\}_{i=1}^{n}$.
 
 We find the Hamiltonian vector field $X_{H_{i}}$ by applying \cref{eq:Hmiltonian_VF} to each $H_{i}$, forming a Hamiltonian system $(\mathbf{R}^2, \omega_0, H_{i})$.
In terms of symbolic computation, the database $\mathcal{X}_{s}(B, \Delta)$ is constructed using the \textit{PoissonGeometry} Python module \cite{EPS1, EPS2}. Specifically, $\mathcal{X}_{s}(B, \Delta)$ is a database of symbolic Hamiltonian vector fields.
In the context of object-oriented programming, each element of $\mathcal{X}_{s}(B, \Delta)$ can be interpreted as an object where $H_{i}$ and $\omega_0$ are attributes represented as a string and a symbolic dictionary, respectively, while $X_{H_{i}}$ is a method that returns a symbolic dictionary.
We define a symbolic dictionary as a variable that contains key-value pairs, where each value is a symbolic element \cite{sympy17}.

\subsection{Numerical database for Hamiltonian vector fiels}\label{subsec:db_numeric}

Next, using the symbolic database $\mathcal{X}_{s}(B, \Delta)$ we construct numerical vector field expressions.
We define the numerical representation of a Hamiltonian vector field $X_{H_{i}}$ by a set of points $\{(u, v) \, | \,  u \text{ and } v\in\mathbf{R}^{2}\}$.
Here $u$ is the base point in the plane and $v$ is a vector with components in the planar $x$ and $y$ directions.
For this, we consider a point cloud, the symbolic expression of a vector field $X_{h}$  and use our Python module \textit{NumericalPoissonGeometry} \cite{EPS2, EPS3}.

\subsubsection{Point Clouds}\label{subsubsec:point_clouds}

\begin{figure}[!htbp]
    \includegraphics[width=1\textwidth]{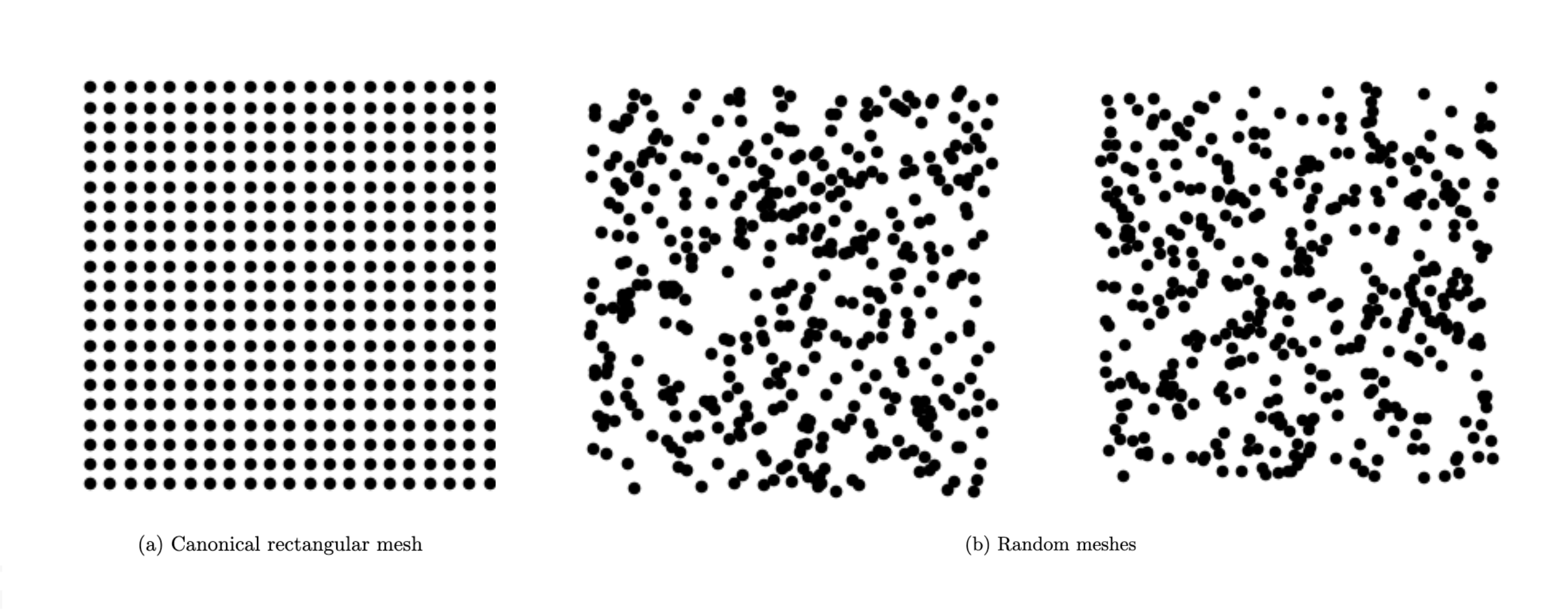}
    \caption{Examples of canonical (4a) and random (4b) point clouds defined in $[-10, 10]^{2}\subset \mathbf{R}^{2}$}
    \label{fig:meshes}
    \centering
\end{figure}

To construct the point meshes, we will consider $50$ distinct point clouds, each uniformly distributed over the domain $[-10, 10]^{2}$, and containing \(21^2\) points.
We include the rectangular lattice, defined as $\{(x,y)| x,y \in [\pm 0, \pm 1, \cdots, \pm 10]\}$ (see \Cref{fig:meshes}a). 
Additionally, consider 49 collections of random point clouds (see \Cref{fig:meshes}b), where any value within the given interval is equally likely to be drawn; that is, the probability density function of the uniform distribution is $p(x)=\frac{x}{b-a}$ anywhere within the interval $[a,b)$ and zero elsewhere. 

All of our point clouds have the same cardinality. 
In principle, larger point clouds could be constructed for greater precision, at greater computational expense. 
Here, this granularity is limited by our available computational resources.

\subsubsection{Numerical database}\label{subsubsec:db_numeric}

We can now produce a diverse range of numerical Hamiltonian systems by evaluating energy functions over a multitude of point clouds. 
Let $\{P_{j} \}_{j=1}^{50}$ be the set of $50$ point clouds constructed above in \cref{subsubsec:point_clouds}.
This selection was made experimentally, bounded by our computational limits. 
This collection of point clouds augments the available training data, aiming to increase the \textit{SymFlux} neural network accuracy for each vector field $X_{H_{i}}$ (see \cref{sec:symflux}).
Evaluating every $H_{i}\in{\rm Ham}(B, \Delta)$ on each point cloud in $\{P_{j}\}_{j=1}^{50}$ (using the Python module \textit{NumericalPoissonGeometry} \cite{EPS3}) we compute a numerical representation $X_{H_{i}}|{P_{j}}$ of the vector field. 

Similarly to \cref{subsec:db_symbolic}, we define a numerical database $\mathcal{X}_{n}(B, \Delta)$ as follows:
$$\mathcal{X}_{n}(B, \Delta) := \{(H_i, \omega_{0_i}, P_{j}, X_{H_i}|{P_j}) \text{ }|\text{ } H_{i}\in{\rm Ham}(B, \Delta), P_{j}\in \{P_{j} \}_{j=1}^{50} \text{ and } \omega_{0} ={\rm d} x\wedge{\rm d}y\}_{i=1}^{n}$$

Within the context of object-oriented programming, and similarly as we explained for $\mathcal{X}_{s}(B, \Delta)$, each element in $\mathcal{X}_{n}(B, \Delta)$ can be interpreted as an object. 
In this light, $\omega_{0}$, $H_{i}$, and $P_{j}$ are object attributes, where $\omega_{0}$ is a symbolic dictionary, $H_{i}$ is a string, and $P_{j}$ is an array. 
Additionally, $X_{H_{i}}$ and $X_{H_{i}}(P_{j})$ are object methods that return a symbolic dictionary and an array, respectively.
Observe that the cardinality of the numerical database $\mathcal{X}_{n}(B, \Delta)$ is $50$ times that of the symbolic database $\mathcal{X}_{s}(B, \Delta)$, as each element in the symbolic database has 50 different numerical representations, one for each point cloud.

\subsection{A data base of visual representations for vector fiels}\label{subsec:db_visual_representations}

Visual representation of vector fields are widely used tools in the analysis of dynamic systems, fluid mechanics, and many scientific disciplines. 
Various methods have been developed to effectively visualize the structure and behavior of these fields \cite{Laramee04}. 
Among the most commonly used approaches are quiver diagrams, which represent the field using arrows. 
These arrows indicate the direction and magnitude of vectors on a discrete grid.
Another method is streamlines, which plot continuous lines that follow the trajectory of particles in the flow. 
These techniques provide an intuitive understanding of the dynamics of the system by illustrating how vectors evolve over space. 
Lastly, heat maps encode the field magnitude using a color scale. This representation highlights variations in intensity without explicitly displaying direction.
These methods have been widely implemented in Python visualization libraries such as Matplotlib, facilitating their use in scientific and engineering applications \cite{Hunter07}.
Inspired by the netCDF format \cite{Rew90} and the aforementioned techniques, we define a visual representation as a mapping  
\begin{equation}\label{eq:visual_representation}
\eta: X_{H}|P_{j} \mapsto \eta(X_{H}|P_{j}) = (\theta_1, \theta_2, \theta_3),    
\end{equation}
where each component $\theta_1$, $\theta_2$, and $\theta_3$ corresponds to a specific visualization technique for numerical representation $X_H|P_{j}$ of the Hamiltonian vector field $X_H$. In our case, $\theta_1$ represents the quiver plot, $\theta_2$ corresponds to the streamplot, and $\theta_3$ denotes the heatmap. 

\begin{figure}[!htbp]
    \includegraphics[height=6.5cm, width=\textwidth]{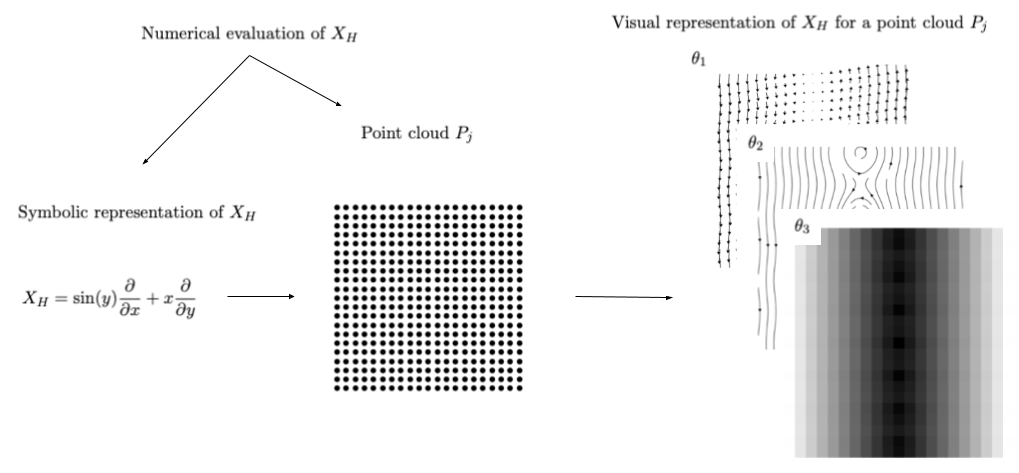}
    \caption{An illustration of how we create visual representations strating from a symbolic expression. In this case, we use as an example the visual representation associated with the Hamiltonian vector field from \cref{eq:PendulumH_x}.}
    \label{fig:visual_representation}
    \centering
\end{figure}

\Cref{fig:visual_representation} shows how a visual representation is generated from a symbolic expression. 
It shows an example of the corresponding visualizations of a Hamiltonian vector field $X_{H}$.
Our approach provides a useful and expressive visual representation of the vector field $X_{H}$. 
All the figures presented in \cref{sec:HamExamples} (\crefrange{fig:HarmonicExa}{fig:Predator-prey}) serve as examples of quiver, streamplot, and heatmap visualizations. 
In addition, the vast majority of models of flow-related data, such as ocean currents and atmospheric circulation, are generated numerically. 
This assumption justifies our above definition of visual representations.
Inspired by data augmentation techniques \cite{Shorten19}, we construct the visual representation of each element in the database of numerical expressions of vector fields, $\mathcal{X}_{n}(B, \Delta)$, defined in \cref{subsec:db_numeric}.  
For this purpose, given a numerical representation $X_{H_{i}}|{P_{j}} \in \mathcal{X}_{n}(B, \Delta)$ of the Hamiltonian vector field $X_{H_{i}}$, we generate its visual representation using the quiver, streamplot, and heatmap methods implemented in Python (see \cref{fig:visual_representation}).
Similarly to \cref{subsec:db_numeric} and \cref{subsec:db_symbolic}, we  define a database $\mathcal{X}(B, \Delta)$ as follows:
$$\mathcal{X}(B, \Delta) := \{(H_i, \omega_{0_i}, \eta(X_{H_{i}}|P_{j})) \text{ }|\text{ } H_{i}\in{\rm Ham}(B, \Delta), P_{j}\in \{P\}_{j=1}^{50} \text{ and } \omega_{0} ={\rm d} x\wedge{\rm d}y\}_{i=1}^{n}$$

Once again, in an object-oriented perespective, each element in $\mathcal{X}(B, \Delta)$ may be interpreted as an object.
The attributes and methods are the same as those in the set $\mathcal{X}_{n}(B, \Delta)$, with the addition of a method $\eta$ that returns an array of arrays representing the visual representation of the vector field $X_{H_{i}}|_{P_{j}}$.
Finally, it is important to note that the cardinality of the data base $\mathcal{X}(B, \Delta)$ depends on the cardinality of the set ${\rm Ham}(B, \Delta)$ multiplied by the number of point clouds used to create the visual representations.
From \cref{tab:basisCard} and \cref{tab:basisTrigoCard}, we highlight the sets ${\rm Ham}(B, \Delta)$ that were feasible to work with based on our available computational resources. The following table presents the cardinality of the set $\mathcal{X}(B, \Delta)$ generated from these ${\rm Ham}(B, \Delta)$ sets.

\begin{table}[ht!]
    \centering
    \begin{tabular}{|c|c|c|c|c|}
        \hline
        Set & ${\rm Ham}(B_{2}, \Delta_3)$ & ${\rm Ham}(B_{2}, \Delta_5)$ & ${\rm Ham}(B_{3}, \Delta_3)$ & ${\rm Ham}(B^{*}_{2}, \Delta_3)$\\
        \hline	
        Cardinality & 12,100 & 160,650 & 984,150 & 984,150 \\
        \hline
        \end{tabular}
    \caption{Cardinality of the database $\mathcal{X}$ for the sets ${\rm Ham}(B_{i}, \Delta_{l})$ for $i \in \{2,3\}$ and $l \in \{3,5\}$.}
    \label{tab:visual_Card}
\end{table}

In summary, each class $X_H$ has $50$ visual representations in the database $\mathcal{X}(B, \Delta)$. 
Increasing the number of point clouds becomes a memory problem that grows linearly. 
For comparison, the sets ${\rm Ham}(B_{3}, \Delta_3)$ and ${\rm Ham}(B^{*}_{2}, \Delta_3)$ contain $19,682$ distinct classes.
This number of classes is smaller than that of the ImageNet database \cite{imagenet09}.

%% file: sec/4_models.tex
\section{\textit{SymFlux}: Methods and Models} \label{sec:symflux}

\subsection{CNN Model}\label{subsec:cnn_model}
CNNs are a class of deep learning models designed to process grid-like data, such as images, by leveraging spatial hierarchies of features. 
They consist of convolutional layers that apply learnable filters to extract local patterns, followed by pooling layers that reduce dimensionality while preserving essential information. 
Due to their efficiency in feature extraction, CNNs have become the standard for tasks such as image classification, object detection, and visual pattern recognition \cite{Lecun98}.

\subsection{LSTM Model}\label{subsec:lstm_model}
LSTM networks are a type of recurrent neural network (RNN) designed to model sequential data by maintaining long-term dependencies. 
Unlike traditional RNNs, LSTMs incorporate a gating mechanism (comprising input, forget, and output gates) that enables them to regulate the flow of information through the network. 
This structure allows LSTMs to effectively capture patterns in natural language processing (NLP) or symbolic sequence generation. 
Their ability to process variable-length input sequences makes them well-suited for tasks that require contextual memory, such as sequence-to-sequence learning \cite{Hochreiter97}.

\subsection{Our hybrid architecture design}\label{subsec:archi_desing}
Image captioning is a task that involves generating descriptive textual information from visual input \cite{vinyals15}.
One approach is to integrate a convolutional neural network (CNN) for the extraction of visual features, along with a long short-term memory network (LSTM) for the generation of descriptive sequences.

In our approach, we consider as input a visual representation $\eta(X_{H}|P_{j})$ associated with a Hamiltonian vector field $X_{H_{i}}$. 
The symbolic expression of the Hamiltonian function $H$ that generates it serves as the description. 
Accordingly, our CNN-LSTM architecture is specifically designed to process data derived from the visual representation of the vector field. 
CNNs extract features from these visual representations, while LSTMs construct the Hamiltonian function that generates the vector field, enabling us to perform symbolic regression (see \cref{fig:symflux}).
We will first explain the tokenization process used to convert the symbolic expressions of Hamiltonian functions into a suitable numerical format. 
Next, we will describe the CNN component of the model, which is responsible for extracting visual features from the representations of the Hamiltonian vector fields. 
Finally, we will present the LSTM component, which processes the extracted features to generate the corresponding Hamiltonian functions, enabling symbolic regression. 

\subsection{Tokenization of Hamiltonian vector fields}\label{subsec:tokenization}
How we tokenize image descriptions from the dataset is similar to how words are tokenized in NLP tasks.
As mentioned in \cref{subsec:archi_desing}, given a visual representation $\eta(X_{H}|P_{j})$of the vector field $X_{H}$, the corresponding description is defined by the Hamiltonian function $H$. 

The objective is that, starting from a visual representation of a flow, our {\it SymFlux} model can find a polynomial $H$ that best describes this motion using a Hamiltonian system.
Since the Hamiltonian functions $H$ in the dataset $\mathcal{X}(B, \Delta)$ are composed of linear combinations of monomials $m_{j}$ of the form $b_{j} x^{h} y^{k}$ (see \cref{subsub:function_polynomial}) or $b_{j} t_{l}$ (see \cref{subsub:function_trigo}), we can interpret the full expression of the Hamiltonian function $H$ as a sentence.
Similarly, each monomial $m_j$ can be regarded as a word within the sentence representation of $H$. 
Here, $m_{j}$ and $-m_{j}$ are different words.

We will now clarify our tokenization process with an example. Consider the $1$-dimensional harmonic oscillator given in \cref{sec:HarmonicExa}.
The associated Hamiltonian function is $H(x,y) = \frac{1}{2}y^2 + x^{2} \in {\rm Ham}(B_{2}, \Delta_{5})$.
Extracting all the monomials of $H$,  we obtain the token set
${\rm tok}_{H} = \{\frac{1}{2}y^2, x^{2}\}$.
From table \ref{tab:basisCard}, we see that the cardinality of the set ${\rm Ham}(B_{2}, \Delta_{5})$ is $3,213$. 
Repeating this process for every $H \in {\rm Ham}(B_{2}, \Delta_{5})$ we obtain $3,213$ token sets.
Let $i \in [0,\dots, \, 3212] \text{ and } H_{i} \in {\rm Ham}(B_{2}, \Delta_{5})$.
Define the set  ${\rm Tok}[{\rm Ham(B_{2}, \Delta_{5})}]$ as the union of all unique such token sets (to avoid repetition).
The collection ${\rm Tok}[{\rm Ham(B_{2}, \Delta_{5})}]$ contains all the monomials that will be used to describe visual representations $\eta(X_{H}|P_{j})$ described above. 

Fix an order on the set ${\rm Tok}[{\rm Ham(B_{2}, \Delta_{5})}]$ and let $N={\rm length}({\rm Tok}[{\rm Ham(B_{2}, \Delta_{5})}])$.
We define a mapping, where each monomial $m_{j}\in {\rm Tok}[{\rm Ham(B_{2}, \Delta_{5})}]$ is assigned $1$ at the $i$-th entry of a length $N$ vector, and zero otherwise.
We thus define a {\it vectorization} mapping, as follows: 

$$m_{i}\in {\rm Tok}[{\rm Ham(B_{2}, \Delta_{5})}] \to (\underbrace{0}_{0-th}, ..., 0, 
 \underbrace{1}_{i-th}, 0, ..., 0, \underbrace{0}_{N-th})\in\mathbf{R}^{N}$$
Given an energy function $f$, we denote its vector representation obtained through the tokenization process as $\mathbf{x}_{f}$
This method allows us to express each visual representation label with a unique vector, facilitating the manipulation and comparison of different representations. 
By assigning a distinct vector to each label, we achieve a precise and efficient mathematical representation that can be utilized in machine learning models for further processing. 
This technique is particularly useful for structuring and organizing large datasets, such as the Hamiltonian vector fields in our study, thus enabling greater flexibility and accuracy.

\subsection{Training}
We work with three CNN models: Xception \cite{Chollet16}, RestNet \cite{he16}, and \textit{SymFlux} (which we developed).
We use four visual representation databases $\mathcal{X}({\rm Ham}(B_{2}, \Delta_{3})) $, $\mathcal{X}({\rm Ham}(B_{2}, \Delta_{5}))$, $\mathcal{X}({\rm Ham}(B_{3}, \Delta_{3}))$, and $\mathcal{X}({\rm Ham}(B_{2}^{*}, \Delta_{3}))$. 
The datasets were split into $75\%$ training and $25\%$ testing. 
\Cref{tab:cnn_train_results} shows the performance of our {\it SymFlux} models over $50$ epochs.
\begin{table}[!htbp]
    \centering
    \begin{tabular}{|c|c|c|}
    \hline
    \,\, Models\,\, & \,\, Train Dataset \,\, & \,\, Accuracy \,\, \\
    \hline
    SymFluxRN    & $\mathcal{X}({\rm Ham}(B_{3}, \Delta_{3}))$ &  $ 86\%$ \\
    SymFluxRN    & $\mathcal{X}({\rm Ham}(B_{2}, \Delta_{5}))$ &  $ 82\%$ \\
    SymFluxRN    & $\mathcal{X}({\rm Ham}(B_{2}, \Delta_{3}))$ & $ 78\%$ \\
    SymFluxRN    & $\mathcal{X}({\rm Ham}(B_{2}^{*}, \Delta_{3}))$ & $ 83\%$ \\
    \hline
    SymFluxX    & $\mathcal{X}({\rm Ham}(B_{3}, \Delta_{3}))$ &  $ 87\%$ \\
    SymFluxX    & $\mathcal{X}({\rm Ham}(B_{2}, \Delta_{5}))$ &  $ 84\%$ \\
    SymFluxX    & $\mathcal{X}({\rm Ham}(B_{2}, \Delta_{3}))$ & $ 76\%$ \\
    SymFluxX    & $\mathcal{X}({\rm Ham}(B_{2}^{*}, \Delta_{3}))$ & $ 83\%$ \\
    \hline
    SymFlux    & $\mathcal{X}({\rm Ham}(B_{3}, \Delta_{3}))$ &  $ 85\%$ \\
    SymFlux    & $\mathcal{X}({\rm Ham}(B_{2}, \Delta_{5}))$ &  $ 82\%$ \\
    SymFlux    & $\mathcal{X}({\rm Ham}(B_{2}, \Delta_{3}))$ & $ 73\%$ \\
    SymFlux    & $\mathcal{X}({\rm Ham}(B_{2}^{*}, \Delta_{3}))$ & $ 78\%$ \\
    \hline
    \end{tabular}
    \caption{Experimental results of our {\it SymFlux} models. 
    Here, SymFluxRN used a ResNet CNN, SymFluxX used an Xception CNN, and \textit{SymFlux} used a CNN developed by us.
    Each model was trained for 50 epochs.}
    \label{tab:cnn_train_results}
\end{table}

To develop the \textit{SymFlux} models, we leveraged the TensorFlow framework \cite{tensorflow15}, employing a learning rate set at $0.0001$ alongside the Adam optimizer \cite{KingBa15}. 
The \textit{SymFlux} model parameters were evaluated iteratively, using a cross-entropy loss metric to assess the prediction accuracy within each completed sentence. 
We used the Keras Tuner \cite{omalley19} module for hyperparameter optimization in the \textit{SymFlux} models.
This led to a performance increase of approximately $5\%$.
The computations involved in the above deep learning tasks were performed on: a laptop computer with an Intel Core i7 processor, equipped with 32 GB of memory RAM, and a single NVIDIA GeForce RTX 2070 GPU, and on UNAM's Miztli supercomputer, with 23 TB of RAM and 16 NVIDIA M2090 GPUs. 
Training cycles took nineteen to twenty-two hours for $50$ epochs, on each {\it SymFlux} model on our laptop machine, and eight to ten hours for the same epochs, on UNAM's Miztli\footnote{\url{http://www.lancad.mx/}} supercomputer.

\subsection{Testing}
Measuring the performance of our {\it SymFlux} models is a complex task, in part because there are no standard, well-established measures for this type of problem.
For this reason, we first tried classical measures used in natural language processing research, such as the Levenshtein or Jaccard measures.

\subsubsection{Levenshtein Measure}\label{subsec:levenshtein}
Recall that the Levenshtein measure, also known as edit distance, quantifies the difference between two sequences by the minimum number of single-character edits (insertions, deletions, or substitutions) required to transform one sequence into the other.
It was originally introduced in the context of error-correcting codes \cite{levenshtein66}.
In the evaluation of testing in deep learning, particularly for tasks involving sequence generation such as natural language processing or DNA sequence analysis, the Levenshtein distance is used to assess the similarity between a generated output sequence and a reference sequence. 
A lower Levenshtein distance indicates higher similarity and thus better performance of the deep learning model in generating accurate sequences.
The Levenshtein distance between two strings $a$ and $b$ is defined as the minimum number of single-character operations insertions, deletions, or substitutions required to transform $a$ into $b$.
Let $d(i, j)$ denote the distance between the prefixes $a = a_1 \ldots a_i$ and $b = b_1 \ldots b_j$. Then:
$$
d(i, j) = \begin{cases}
\max(i, j) & \text{if } \min(i, j) = 0, \\
\min \begin{cases}
d(i-1, j) + 1, \\
d(i, j-1) + 1, \\
d(i-1, j-1) + \delta(a_i, b_j)
\end{cases} & \text{otherwise},
\end{cases}
$$

where $\delta(a_i, b_j) = 0$ if $a_i = b_j$, and $1$ otherwise.

In the Levenshtein measure, the order of the labels is important. 
For example, consider the labels $h_{1} = x^{3}+xy^{2}+x^{2}y+y^{3}$ and $h_{2} = y^{3} +xy^{2}+x^{2}y+x^{3}$, where each monomial is a word.
Therefore, the Levenshtein distance between $h_1$ and $h_2$ is not zero, and it is thus not suitable for our purposes.

\subsubsection{Jaccard distance}\label{subsec:jaccard}
The Jaccard index, of a pair of sets $A=\{a_1, \ldots, a_i\}$ and $B=\{b_1, \ldots, b_j\}$, is calculated as the ratio of the size of their intersection to the size of their union: $J(A, B) = \frac{|A \cap B|}{|A \cup B|}$. 
The Jaccard distance is a measure of dissimilarity between two finite sets, defined as \cite{jaccard1901etude}: 
 $$D_J(A, B) = 1 - J(A, B) = \frac{|A \cup B| - |A \cap B|}{|A \cup B|}$$ 
In the context of evaluating testing in deep learning, the Jaccard distance is often used to compare the similarity of sets, such as the set of predicted labels versus the set of true labels in multi-label classification, or the set of detected objects in an image compared to the ground truth objects in object detection. 
A lower Jaccard distance indicates greater similarity between the sets being compared.

The labels $h_1$ and $h_2$ shown in \cref{subsec:levenshtein} are correctly measured in the Jaccard metric, regardless of the order. However, observe that it only measures the matches of the elements whose tokenization has value $1$. 
Thereby disregarding all of the positive predictions that match elements with tokenization value $0$.
This penalizes the learning process and does not properly reflect the accuracy of the prediction.

\subsubsection{Our proposed metric}
As we explained above, the previous two metrics did not work correctly in our setup.
Inspired by the Jaccard metric (\cref{subsec:jaccard}) we designed an \textit{ad hoc} metric for our task. 
\begin{definition}\label{def:our_distance}
Let $f, g\in {\rm Ham}(B, \Delta)$ (see \cref{subsubsec:hamiltonian_set}), and $\mathbf{x}_{f}=(x_0, \cdots, x_n), \mathbf{y}_{g}=(y_0, \cdots, y_n)\in\mathbf{R}^{N}$ their respective vector representations of the tokenization process described in \cref{subsec:tokenization} with $N={\rm length}({\rm Tok}[{\rm Ham(B, \Delta)}])$.
Define the distance of Hamiltonian functions as 
\begin{equation}\label{eqn:our-dist}
    d(f,g) = d(\mathbf{x}_{f}, \mathbf{y}_{g}) : = \sqrt{\sum_{i=0}^{n} (x_i - y_i)^2}
\end{equation}
 
\end{definition}
Our metric takes the tokenized vector $\mathbf{x}_{f}$, associated with the true energy function, and the vector $\mathbf{y}_{g}$, associated with the predicted function, and computes the standard Euclidean distance between them.
With our distance $d(f,g)$, values close to zero indicate that the two compared vectors are very similar or nearly identical in the vector space $\mathbf{R}^N$.
In this way, it is possible to compare all existing matches (including all $0$ and $1$ values in the tokenized vector) between two labels. 
Therefore, our method does a better job of recovering positive predictions.
\Cref{tab:cnn_test_results} shows the result obtained with the distance measure expressed in \cref{def:our_distance} with the test data set. 
\begin{table}[!htbp]
    \centering
    \begin{tabular}{|c|c|c|}
    \hline
    \,\, Models\,\, & \,\, Test Dataset \,\, & \,\, Accuracy \,\, \\
    \hline
    SymFluxRN    & $\mathcal{X}({\rm Ham}(B_{3}, \Delta_{3}))$ &  $ 88\%$ \\
    SymFluxRN    & $\mathcal{X}({\rm Ham}(B_{2}, \Delta_{5}))$ &  $ 83\%$ \\
    SymFluxRN    & $\mathcal{X}({\rm Ham}(B_{2}, \Delta_{3}))$ & $ 81\%$ \\
    SymFluxRN    & $\mathcal{X}({\rm Ham}(B_{2}^{*}, \Delta_{3}))$ & $ 86\%$ \\
    \hline 
    SymFluxX    & $\mathcal{X}({\rm Ham}(B_{3}, \Delta_{3}))$ &  $ 88\%$ \\
    SymFluxX    & $\mathcal{X}({\rm Ham}(B_{2}, \Delta_{5}))$ &  $ 86\%$ \\
    SymFluxX    & $\mathcal{X}({\rm Ham}(B_{2}, \Delta_{3}))$ & $ 79\%$ \\
    SymFluxX    & $\mathcal{X}({\rm Ham}(B_{2}^{*}, \Delta_{3}))$ & $ 85\%$ \\
    \hline
    SymFlux    & $\mathcal{X}({\rm Ham}(B_{3}, \Delta_{3}))$ &  $ 87\%$ \\
    SymFlux    & $\mathcal{X}({\rm Ham}(B_{2}, \Delta_{5}))$ &  $ 85\%$ \\
    SymFlux    & $\mathcal{X}({\rm Ham}(B_{2}, \Delta_{3}))$ & $ 77\%$ \\
    SymFlux    & $\mathcal{X}({\rm Ham}(B_{2}^{*}, \Delta_{3}))$ & $ 81\%$ \\
    \hline
    \end{tabular}
    \caption{Experimental results of our {\it SymFlux} models in the test dataset with our metric (defined in \cref{def:our_distance}). 
    Remember that SymFluxRN used a ResNet CNN, SymFluxX used an Xception CNN, and \textit{SymFlux} used a CNN was developed by us.}
    \label{tab:cnn_test_results}
\end{table}

In \Cref{tab:predictions} we present the best predictions obtained by our \textit{SymFlux} models (SymfluxX, SymfluxRN, or Symflux) for the examples discussed in \cref{sec:Ham}.
\begin{table}[!htbp]
    \centering
    \begin{tabular}{|l|c|c|}
    \hline
    \,\, Example \,\, & \,\, Ground truth \,\, & \,\, Best Prediction \,\, \\
    \hline
    ($\S$\ref{sec:HarmonicExa}) 1-dim. harmonic oscillator & $H(x,y) = \frac{1}{2}(y^2 + x^{2})$ & $H(x,y) = \frac{1}{2}y^2 + \frac{1}{2}x^{2}$ \\
    \hline 
    ($\S$\ref{sec:PendulumExa}) Mathematical pendulum & $H(x,y) = \frac{1}{2}x^2 + \cos (y)$ & $H(x,y) = x^2 + \cos(y)$ \\
    \hline 
    ($\S$\ref{sec:Lotka-Volterra-ex}) Lotka–Volterra system  & $H(x,y) = x \ln (x) + y \ln (y)$ & $H(x,y) = x^{2} - xy - cos(y)$ \\
    & $ - 1.1x - 1.1y - 0.1xy$ \, & \,\,  $+ y^{2} + y $ \\
    \hline 
    ($\S$\ref{sec:SISHamiltonian}) Hamiltonian SIS Model  & $H(x,y) = xy(1-x) + \frac{1}{y}$ & $H(x,y) = \frac{1}{2}xy+ xy^{2} - x^{2}$ \\
    \hline
    \end{tabular}
    \caption{We show the best predictions obtained by the \textit{SymFlux} model for the examples shown in \cref{sec:Ham}.}
    \label{tab:predictions}
\end{table}

From \cref{tab:predictions}, the following observations can be made: First, the prediction for the one-dimensional harmonic oscillator case is perfectly accurate. 
Second, in the case of the mathematical pendulum, the predicted Hamiltonian function $H(x,y) = x^2 + \cos (y)$, differs from the original function $H(x,y) = \frac{1}{2}x^2 + \cos (y)$, only by a constant. 
As a result, the corresponding vector fields are almost identical: $X_{H} = \sin(y)\frac{\partial}{\partial x} + x\frac{\partial}{\partial y}$ corresponds to the original Hamiltonian function and $X_{H} = \sin(y)\frac{\partial}{\partial x} + 2x\frac{\partial}{\partial y}$ to the predicted Hamiltonian function.
Our model does not detect this difference, as both energy functions produce identical vector field representations (see \cref{subsec:db_visual_representations}).

\begin{figure}[!htbp]
    \includegraphics[width=0.97\textwidth]{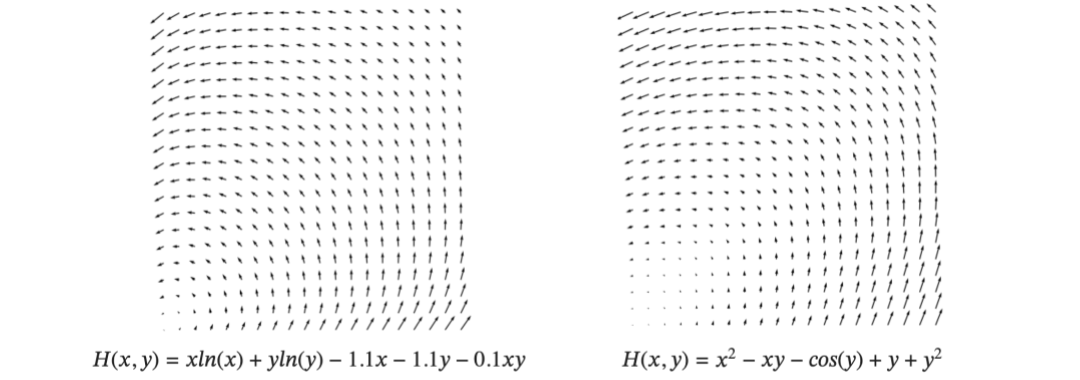}
    \caption{Visual representation of the vector field: $(0.1x - log(y) + 0.1)\frac{\partial}{\partial x} + (log(x) -0.1y - 0.1)\frac{\partial}{\partial x}$ for the original energy function $H(x,y) = x \ln (x) + y \ln (y)  - 1.1x - 1.1y - 0.1xy $ (left) and $(x - 2y - sin(y) - 1)\frac{\partial}{\partial x} + (2x - y)\frac{\partial}{\partial y}$ for the predicted energy function $H(x,y) = x^{2} - xy - cos(y) + y^{2} + y$ (right) for the Lotka–Volterra system.}
    \label{fig:prediction_comparision}
    \centering
\end{figure}

In the Lotka--Volterra system, although the Hamiltonian function was not accurately recovered, the representation of the vector field resulting closely resembles the original, as illustrated in \cref{fig:prediction_comparision}.
Observe that, as the sets of Hamiltonian functions that we used for training do not include logarithmic functions, the current \textit{SymFlux} systems cannot be expected to reproduce the ground-truth expression for $H(x,y)$ in the Lotka--Volterra system.

For the Hamiltonian SIS model, both the predicted Hamiltonian function and the associated vector field differ significantly from the originals, also indicating an incorrect prediction.
As in the Lotka--Volterra case, here the Hamiltonian function involves an expression that was not part of the traning data, namely $1/y$. 
So again, for this case, it would not be expected to recover the exact expression.

In order for systems like the Lotka--Volterra or Hamiltonian SIS model to be accurately predicted, we would need to train all the \textit{SymFlux} systems again starting with a basis of Hamiltonian functions including logarithmic functions and multiplicative inverses. 
This, however, is currently beyond our computational resources, as we explained in the construction of the Hamiltonian functions above. 

For those interested in conducting experiments and comparing different architectures, our code repository is available \href{https://github.com/appliedgeometry/symflux}{here}. 

%% file: sec/5_conclusions.tex
\section{Conclusions} \label{sec:conclusions}

As we mentioned in the introduction, the Hamiltonization problem has a local theoretical solution. 
Here, we built an artificial intelligence system that automates the processing of Hamiltonizing vector fields, proposing potential functions whose Hamiltonian vector fields best approximate the initial system.
We developed a novel family of hybrid CNN+LSTM systems that perform symbolic regression and prediction for Hamiltonian vector fields with accuracies of up to 89\% . 
In the process, we create new (synthetic) databases of Hamiltonian vector fields, represented symbolically $\mathcal{X}_{s}$ ($\S$ \ref{subsec:db_symbolic}), numerically $\mathcal{X}_{n}$ ($\S$ \ref{subsec:db_numeric}), and visually $\mathcal{X}$ ($\S$ \ref{subsec:db_visual_representations}).
Our Hamiltonian functions include combinations of polynomial and trigonometric expressions ($\S$ \ref{subsec:functions}). 
Specific examples in which our \textit{SymFlux} systems perform well or fail (when the training data do not include the target functions) are shown in \cref{tab:predictions}.

Considerable learning gains were achieved by augmenting the visual representation data sets (see \cref{tab:cnn_train_results}), leading to an increase in accuracy from $75\%$ to $89\%$.
Therefore, we conclude that our model's learning accuracy increases with the number of visual representations used in training.

The visual representation introduced in \cref{sec:DB} may be scaled as follows: more visualization techniques $\theta_{n}$ could be added to \cref{eq:visual_representation}.
Similarly, the CNN architecture used in the \textit{SymFlux} model (see \cref{sec:symflux}) could also be scaled, hoping to achieve performance gains.
Potential improvements to our approach are possible given more computational power, enabling the processing of larger and more expressive training sets of Hamiltonian vector fields.
Moreover, we chose the LSTM architecture in order to have a clear understanding of each step of the symbolic generation process. 
Future work could include substituting the LSTM module for a Transformer architecture.

%% file: sec/A1_appendix.tex
\section{Data Sheet}\label{app:A}
In this Appendix, we include a Datasheet for the data bases used in our \textit{SymFlux} methods, following the guidelines laid out by Gebru et al. \cite{Gebru21}.
(We did not include answers to questions that are not applicable to our work due to their theoretical and synthetic nature.)  

\definecolor{darkblue}{RGB}{46,25, 110}

\newcommand{\dssectionheader}[1]{%
   \noindent\framebox[\columnwidth]{%
      {\fontfamily{phv}\selectfont \textbf{\textcolor{darkblue}{#1}}}
   }
}

\newcommand{\dsquestion}[1]{%
    {\noindent \fontfamily{phv}\selectfont \textcolor{darkblue}{\textbf{#1}}}
}

\newcommand{\dsquestionex}[2]{%
    {\noindent \fontfamily{phv}\selectfont \textcolor{darkblue}{\textbf{#1} #2}}
}

\newcommand{\dsanswer}[1]{%
   {\noindent #1 \medskip}
}

\begin{multicols}{2}


\small{


\dssectionheader{Motivation}

\dsquestionex{For what purpose was the dataset created?}
{
}
\dsanswer{
Our \textit{SymFlux} databases are the first to include $2$-dimensional vector fields. 
We provide a new set of methods that visually describe $2$-dimensional vector fields.
}

\dsquestion{Who created this dataset (e.g., which team, research group) and on behalf of which entity (e.g., company, institution, organization)?}
\dsanswer{
P. Suárez-Serrato and M. Evangelista-Alvarado created the \textit{SymFlux} database and methods for visualizing two-dimensional Hamiltonian vector fields shown in this article.
In this publication, PSS is affiliated with UCSB and UNAM (National Autonomous University of Mexico), and MAE with the University Rosario Castellanos.
}

\dsquestionex{Who funded the creation of the dataset?}
{
}
\dsanswer{
DGTIC internal UNAM grant to Pablo Suárez-Serrato and CONAHCYT through a PhD scholarship awarded to Miguel Evangelista-Alvarado.
}

\dsquestion{Any other comments?}
\dsanswer{
The \textit{SymFlux} database is a synthetic database of visualizations of two-dimensional Hamiltonian vector fields.
The Hamiltonian functions used to generate the Hamiltonian vector fields are truncated polynomials up to a certain power.
}

\bigskip
\dssectionheader{Composition}

\dsquestionex{What do the instances that comprise the dataset represent (e.g., documents, photos, people, countries)?} 
{
}
\dsanswer{
These are synthetic databases that include functions, Hamiltonian vector fields, and their two-dimensional visualizations, where the associated Hamiltonian function is a polynomial.
Section \ref{sec:DB} explains the methods for generating these visualizations, and the source code is available ( \url{https://github.com/appliedgeometry/symflux}).
}

\dsquestionex{
Does the dataset contain all possible instances or is it a sample (not necessarily random) of instances from a larger set?}
{
}
\dsanswer{
The \textit{SymFlux} database is a sample due to the way it was created. The methods shown in this paper to create a \textit{SymFlux} database require a finite seed set of Hamiltonian functions (which are polynomials in this case), a finite discrete set of coefficients, and a finite set of point clouds. This process determines the cardinality of the \textit{SymFlux} database, as explained in Section \ref{sec:DB}.
}

\dsquestionex{What data does each instance consist of? “Raw” data (e.g., unprocessed text or images) or features?}
{
}
\dsanswer{
Each element of the \textit{SymFlux} database is a visual representation of a two-dimensional Hamiltonian vector field within $[-10, 10]^{2} \in \mathbf{R}^{2}$.

The visual representation is generated as follows: Fix a Poisson bivector $\pi$ and a polynomial function $H:\mathbf{R}^{2} \to\mathbf{R}$. With Python Poisson geometry methods \cite{EPS1}, we compute the Hamiltonian vector field $X_{H}$ in the symbolic sense. We numerically evaluate the vector field $X_{H}$ over a point cloud (previously selected from a fixed set of meshes) contained in $[-10, 10]^{2}$ with Python Numerical Poisson Geometry methods \cite{EPS2}; finally generate the image with Python code.
}

\dsquestion{Is there a label or target associated with each instance?}
\dsanswer{
The label associated with each visual representation of the two-dimensional Hamiltonian vector field $X_{H}$ is the Hamiltonian function $H$.
}

\dsquestionex{Is the dataset self-contained, or does it link to or otherwise rely on external resources (e.g., websites, tweets, other datasets)?}
{
}
\dsanswer{

The software versions used are Python 3.11.6, PoissonGeometry 1.0.2, NumericalPoissonGeometry 1.1.1, and Matplotlib 3.8. The \href{https://github.com/appliedgeometry/symflux}{\textit{SymFlux} repository} contains all the methods used in this work.
}

\dsquestionex{Does the dataset relate to people?}
{
}
\dsanswer{
Our databases and the methods presented in this work do not relate to people.
}

\bigskip
\dssectionheader{Collection Process}

\dsquestionex{How was the data associated with each instance acquired?}
{
}
\dsanswer{
The images that are part of the \textit{SymFlux} database were generated synthetically (with Python code) for academic purposes.
}

\dsquestionex{What mechanisms or procedures were used to collect the data (e.g., hardware apparatus or sensor, manual human curation, software program, software API)?}
{
}
\dsanswer{
The visual representations of $X_{H}$ were created with methods developed by us,
explained in detail in section \ref{sec:DB}, and the source code 
can be seen in the \href{https://github.com/appliedgeometry/symflux}{\textit{SymFlux} repository}.
}

\dsquestion{Who was involved in the data collection process (e.g., students, crowdworkers, contractors) and how were they compensated (e.g., how much were crowdworkers paid)?}
\dsanswer{
Only the authors of this paper participated in the methods for developing the \textit{SymFlux} database.
}

\dsquestionex{Over what timeframe was the data collected? Does this timeframe match the creation timeframe of the data associated with the instances (e.g., recent crawl of old news articles)?}
{
}
\dsanswer{
The database and its methods were developed between 2023 and 2025.
As a synthetic database, anyone can recreate it at any time with expected comparable results. 
}

\bigskip
\dssectionheader{Preprocessing/cleaning/labeling}

\dsquestionex{Was any preprocessing, cleaning, or labeling of the data done (e.g., discretization or bucketing, tokenization, part-of-speech tagging, SIFT feature extraction, removal of instances, processing of missing values)?}
{
}
\dsanswer{
A tokenization of the label set (symbolic expressions of the Hamiltonian functions) was performed to train with a CNN-LSTM model (for details, see section \ref{sec:DB}).
}

\bigskip
\dssectionheader{Uses}

\dsquestionex{Has the dataset been used for any tasks already?}
{
}
\dsanswer{
The \textit{SymFlux} database was developed alongside the \textit{SymFlux} model presented here. 
These were explicitly developed to automatically solve the Hamiltonization problem, as explained in \cref{sec:Ham}. 
Our \textit{SymFlux} models perform symbolic inference of two-dimensional Hamiltonian vector fields.
}

\dsquestionex{Is there a repository that links to any or all papers or systems that use the dataset?}
{
}
\dsanswer{
The \textit{SymFlux} repository: \url{https://github.com/appliedgeometry/symflux}.
}

\dsquestion{What (other) tasks could the dataset be used for?}
\dsanswer{
Our datasets could be used in the solution of problems that involve two-dimensional vector fields.
}

\bigskip
\dssectionheader{Distribution}

\dsquestionex{How will the dataset will be distributed (e.g., tarball on website, API, GitHub)}
{
}
\dsanswer{
The methods for generating the \textit{SymFlux} database can be found in our Github repository: \url{https://github.com/appliedgeometry/symflux}
}

\dsquestion{When will the dataset be distributed?}
\dsanswer{
The \textit{SymFlux} dataset and its methods will be distributed starting from the online release and publication date of this work.
}

\dsquestionex{Will the dataset be distributed under a copyright or other intellectual property (IP) license, and/or under applicable terms of use (ToU)?}
{
}
\dsanswer{
The \textit{SymFlux} dataset and methods are distributed under the MIT License, https://opensource.org/licenses/MIT.
}

\bigskip
\dssectionheader{Maintenance}

\dsquestion{Who will be supporting, hosting, or maintaining the dataset?}
\dsanswer{
Miguel Evangelista-Alvarado. 
}

\dsquestion{How can the owner/curator/manager of the dataset be contacted (e.g., email address)?}
\dsanswer{
We recommend that users contact the administrators through an issue on the \textit{SymFlux} repository: \url{https://github.com/appliedgeometry/symflux}.
}

\dsquestionex{Will the dataset be updated (e.g., to correct labeling errors, add new instances, delete instances)?}
{
}
\dsanswer{
The \textit{SymFlux} dataset and methods will be updated if and when a user opens a meritorious support ticket. 
Changes will be communicated through the \textit{SymFlux} repository on Github.
}

\dsquestionex{If others want to extend/augment/build on/contribute to the dataset, is there a mechanism for them to do so?}
{
}
\dsanswer{
User that wish to build on our work may make a fork of the \textit{SymFlux} repository, respecting the MIT license and the directions in the \textit{SymFlux} repository. 
}

}
\end{multicols}